# Enhancing Metacognitive AI: Knowledge-Graph Population with Graph-Theoretic LLM Enrichment

**Deniz Askin[1], Gal Hadar[2], Brendan Conway-Smith[1]**
[1]Department of Cognitive Science, Carleton University
[2]Faculty of Computer and Information Science, Ben-Gurion University of the Negev

## Abstract

Metacognition—the ability to monitor one's own knowledge state, spot gaps, and autonomously fill them—remains largely absent from modern AI. Here, we present MetaKGEnrich, a fully automated pipeline that endows large language model (LLM) applications with self-directed knowledge repair. The system (i) builds knowledge graphs from a seed query, (ii) detects sparse regions via seven graph metrics, (iii) has GPT-4o generate targeted questions, (iv) retrieves web evidence with Tavily and ingests it into Neo4j, and (v) re-answers the query with GraphRAG for GPT-4 to evaluate improvement. Tested on 30 queries from each of three widely-used datasets: Google Research Natural Questions, MS MARCO, and Hot-potQA. MetaKGEnrich improved answer quality in 80% of HotpotQA questions, 87% of Google Research Natural Questions and 83% of MS MARCO questions, while preserving well-supported regions. This proof of concept demonstrates how topological self-diagnosis plus targeted retrieval can advance AI toward humanlike metacognitive learning.

## Introduction

Humans rely on metacognition to recognize when they do not know and to initiate strategies—such as asking questions or seeking information—to fill those gaps (Flavell 1979; Metcalfe and Shimamura 2009). Contemporary Al systems, despite fluent generation and strong retrieval, still lack this metacognitive capacity. They treat knowledge as static, depend on human intervention to resolve knowledge gaps, and often fail on out-of-scope queries. Addressing this limitation is crucial for building scalable, trustworthy, and self-improving AI.

To overcome these limitations and move toward more autonomous, self-learning, intelligent systems, we introduce Meta-KG-Enrich (MKGE), a pipeline that integrates three key components:

1. **Graph-theoretic sparsity detection.** Seven met-rics—clique membership, isolates (non-clique), clustering coefficient, degree centrality, betweenness, component diameter, and Louvain community size—quantify specific metacognitive shortcomings (Table 1). A node scoring at or below the median for a given metric is flagged as *sparse*.
2. **LLM-driven question generation.** LLM receives sparse nodes and their associated cognitive label (e.g. "bridge missing" for low betweenness) and formulates concise factual questions aimed at raising the metric score.
3. **Automated retrieval and KG update.** Each question triggers a web search through Tavily API, we embed the first snippet with sentence-embedding `all-MiniLM-L6-v2` and push into a Neo4j KG. On average, ~12 new documents are added per user query (one per successful question), while the initial seed KG uses exactly one snippet.

For evaluating the pipeline, we tested MetaKGEnrich on 30 questions from each of the Natural Questions, MS MARCO, and HotpotQA datasets, measuring answer quality before and after enrichment using GraphRAG and an LLM judge. Sparse-node questioning consistently improved answer quality across datasets: 80% on HotpotQA, 87% improvement on Natural Questions, and 83% on MS MARCO.

MetaKGEnrich converts graph topology into metacognitive signals that drive autonomous web exploration and knowledge repair-advancing toward AI systems that not only answer questions, but know when they need to learn more and how to go about.

## Related Work

A substantial body of work has addressed knowledge graph (KG) enrichment, from early rule-based pipelines for triple extraction, to distantly supervised relation mining, to more recent approaches using language models for link prediction, all contributing to improved KG coverage and downstream QA performance (Masoud and Sowa 2021)). However, most existing systems (i) depend on predefined extraction rules or sampled negative edges, and (ii) lack a principled way to identify where enrichment is most needed. Our approach differs in three key ways: (a) it detects sparse regions using seven graph-theoretic metrics; (b) it prompts an LLM to generate metric-aware questions that directly target those gaps; and (c) it closes the loop by automatically retrieving web evidence and updating the KG-independent of any specific query.

To date, no work has integrated sparsity detection, targeted question generation, retrieval, and re-insertion into a fully automated, end-to-end pipeline. The following sections highlight the key strands of prior research that MetaKGEnrich builds upon and extends.

**Context and novelty.** A rich body of research tackles knowledge-graph enrichment: early rule-based pipelines for triple extraction, distantly-supervised relation mining, and more recent language-model–assisted link prediction have all improved KG coverage and downstream QA accuracy (Masoud and Sowa 2021). [1] Yet these systems typically (i) rely on pre-defined extraction patterns or sampled negative edges, and (ii) lack a mechanism to diagnose where enrichment is most needed. Our formulation differs: we **(a)** detect sparse graph regions via seven graph-theoretic metrics, **(b)** let an LLM pose metric-aware questions that target those blind spots, and **(c)** populate the KG with web evidence in a fully automated loop, independent of any single query. To our knowledge, no prior work couples sparsity detection, question generation, retrieval, and re-insertion in an end-to-end fashion.

**Retrieval-Augmented Generation (RAG)** Lewis et al. (2020) first unified parametric LMs with non-parametric retrieval, reducing hallucination by drawing external evidence. Follow-ups study vector indices, hybrid BM25+dense search, and continual refresh (e.g. LAVA) for long-tail knowledge (Izacard and Petroni 2022; Zhang and Gillick 2022). Integration with graph databases spawned GraphRAG, which exploits explicit entity-relation context in retrieval (Hunger 2024a; Engineering 2024a). Our pipeline inherits RAG's evidence-grounding but actively curates the KG itself, turning retrieval into a self-improving loop.

**Iterative self-feedback Self-Refine** iteratively critiques and rewrites its own outputs without additional training (Madaan et al. 2023). **Reflexion** externalises feedback as explicit "reflection tokens" stored in memory to guide future actions (Shinn, Yao, and Narasimhan 2023). We adopt the high-level idea of self-repair but make the feedback structural: graph metrics quantify gaps, and the LLM supplies *questions* (not edits) whose answers materially enlarge the graph.

**Chain-of-Thought prompting** Reasoning traces elicited via chain-of-thought (CoT) substantially boost arithmetic and commonsense accuracy (Wei et al. 2022). Subsequent work explores few-shot CoT, rationale distillation, and program-of-thought execution. Our metric-guided questions act as *externalised* CoT: instead of inner monologues, the model converts uncertainty into explicit sub-questions whose answers persist as KG nodes, enabling later reuse and auditability.

**Automatic KG construction and enrichment** Comprehensive surveys document text-to-KG pipelines, joint entity–relation extraction, and KG completion via graph embeddings (Masoud and Sowa 2021). Embedding-based completion excels at link prediction but is blind to topical sparsity; graph-attention Populators inject new entities but still rely on fixed candidate pools (Xu, Wu, and Tang 2025). In contrast, MKGE continuously discovers missing content through metric-conditioned web queries, sidestepping fixed candidate constraints.

[1] Masoud and Sowa (2021) survey 300+ enrichment pipelines covering IE, entity linking, and embedding-based completion.

**Graph-metric-based active learning** Graph centrality has guided active sampling in semi-supervised node classification (Xu, Wu, and Tang 2025) and graph-level regression (Chen and Liu 2024), showing that low-degree or bridge nodes carry disproportionate information value. We extend this principle from labelling to knowledge ingestion: low-metric nodes trigger enrichment rather than annotation.

**Community detection for sparsity diagnostics** Louvain's fast unfolding remains a de-facto standard for scalable community detection (Blondel et al. 2008; Yang and Leskovec 2023). We reinterpret small community size as a cognitive signal of poor inter-module transfer and use it to seed cross-cluster questions.

**Summary of differences** Unlike prior RAG or KG-completion systems that passively retrieve or predict edges, our pipeline closes the loop: topological diagnostics → LLM question generation → targeted retrieval → KG insertion → re-evaluation. This yields an automated, query-agnostic path to KG health that unifies ideas from RAG, self-feedback, and graph-based sampling in a single metacognitive controller.

# Methodology

## Pipeline Overview

To enable self-directed knowledge repair, MetaKGEnrich combines graph analysis, large language models, and retrieval in a structured five-stage loop. The pipeline begins with a minimal seed knowledge graph and iteratively expands it by identifying informational gaps, generating targeted questions, retrieving external evidence, and updating the graph. This section describes the architecture and key components of the system, beginning with a high-level overview.

Figure 1 sketches the five–stage loop; Algorithm 1 provides the corresponding pseudocode. The stages are as follows:

1. **Stage 1 – KG seeding.** A single web query is issued via the Tavily API. We retain only the top-ranked snippet and ingest it as one 500-token document into Neo4j using the GraphRAG pipeline.[2] Retaining just the first snippet allows us to assess whether a given question is easily answerable (e.g., binary, factual) or whether it requires supplementary information—critical for isolating the cases where enrichment makes a measurable difference.
2. **Stage 2 – Sparse-node detection.** We compute seven graph-theoretic metrics—clique membership, non-clique status, clustering coefficient, degree, betweenness, component diameter, and Louvain community size—using NetworkX 3.5 (Developers 2024) A node is labeled sparse if it scores at or below the median for a given metric, yielding an adaptive, data-driven threshold.

[2] Neo4j 5.16; Python driver 5.18 (Engineering 2024b)

3. **Stage 3 – Question generation.** GPT-4o (OpenAI, 2025) receives up to 50 sparse node IDs, a 160-character preview of each, and their associated metric labels. It generates five fact-seeking questions per metric (temperature 0), producing a total of 7 × 5 = 35 enrichment questions per user query.
4. **Stage 4 - Retrieval Ingestion.** For each question, we retain at most the first web result. Retrieved content is chunked into 500-character spans, embedded using all-MiniLM-L6-v2 (Reimers Gurevych, 2021), and integrated into the Neo4j graph.
5. **Stage 5 – Evaluation.** GraphRAG retrieves the top five semantically closest KG chunks before and after enrichment. These are fed to GPT-4 for answer generation. A separate GPT-4 judge labels each answer pair as Improved or Not Improved.

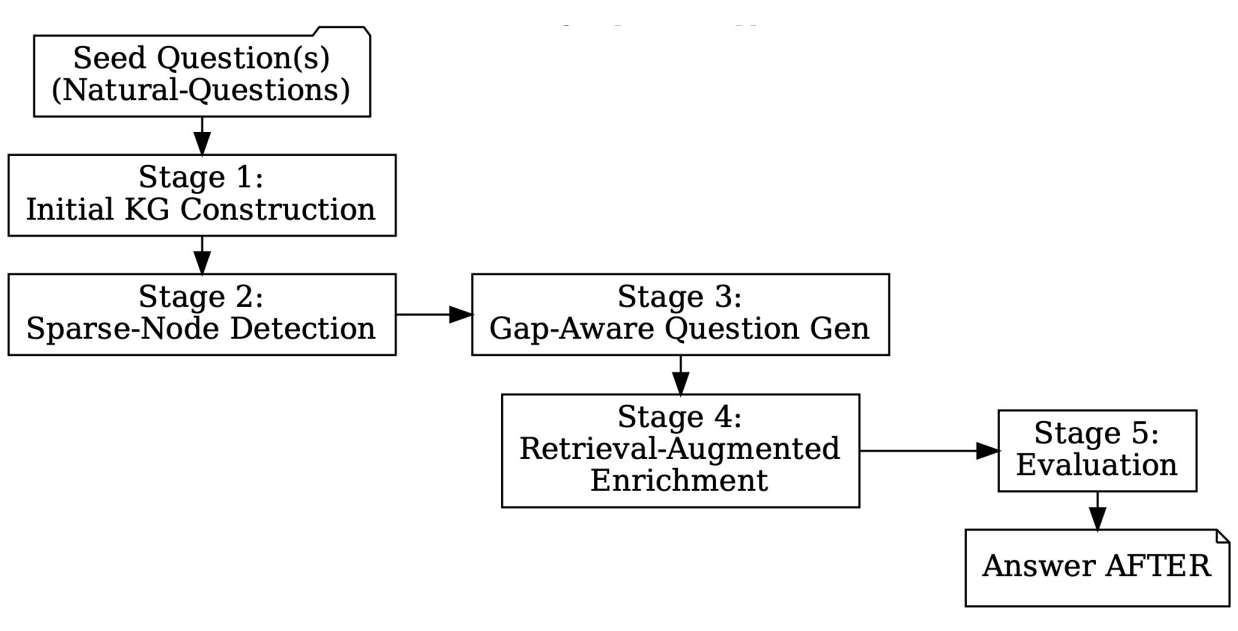

Figure 1: MKGE architecture. Metric-guided questioning expands sparse regions and re-evaluates answers.

**Core software modules. Neo4j** for graph storage (Engineering 2024b); **Tavily** search API for web snippets (AI 2025); **GraphRAG** retriever on Neo4j (Hunger 2024b); **FAISS** for local cosine indexing of chunk embeddings (Johnson, Douze, and Je´gou 2021); **NetworkX** for metric computation (Developers 2024).

**Algorithm 1:** MKGE

**Input:** Question set Q, metric registry M
**foreach** $q \in$ Q **do**
  KG ← INGEST(TAVILYSEARCH($q$));
  $\text{ans}_{\text{before}}$ ←GRAPHRAG($q$,KG);
  **foreach** $m \in$ M **do**
    sparse ← DETECTSPARSE($KG, m$);
    $Q_m$ ←LLMQUESTIONGEN($sparse, m$);
    docs ← TAVILYSEARCH($Q_m$);
    KG ← INGEST(docs);
  $\text{ans}_{\text{after}}$ ←GRAPHRAG($q$,KG);
  verdict ←JUDGELLM($\text{ans}_{\text{before}}$,$\text{ans}_{\text{after}}$);

```
for idx, q in enumerate(questions, 1):
   delete_all_indexes(driver); _clear_graph(
       driver)

   # ---- Stage 1: seed KG ----
   seed_txt = tavily.search(q, max_results
       =3)["results"][0]["content"]
   build_kg_from_docs(driver, llm, embedder,
        [seed_txt], INDEX)

   ans_before = graphrag_answers(driver,
       embedder, llm, INDEX, [q], "BEFORE")
       [0]

   # ---- Stage 2-4: metric-guided
       enrichment ----
   for mkey, cognitive, shortcoming in
       EMBEDDED_MAPPING:
      run_metric_cycle(
         driver, tavily, llm, embedder,
         metric_key=mkey,
         shortcoming=shortcoming,
         blindspot_topic=q,
         index_name=INDEX
      )

   # ---- Stage 5: evaluation ----
   ans_after = graphrag_answers(driver,
       embedder, llm, INDEX, [q], "AFTER")
       [0]
   verdict = evaluate_before_after(eval_llm,
                                [q], [ans_before
                                   ], [ans_after
                                   ])[0]
```

Listing 1: Excerpt from our code `Meta-KG-Enrich.py` illustrating two GraphRAG executions: one using the initial knowledge graph generated via Tavily search, and the other using the enriched knowledge graph produced by MKGE.

## Graph Metrics and Sparsity Quantification

We compute seven metrics with NetworkX: Clique membership, Non-Clique (isolates), Clustering coefficient, Degree centrality, Betweenness, Component Diameter, and Louvain community size. A node whose score $\leq$ median for that metric is deemed sparse. Median thresholding adapts across heterogeneous graphs and avoids hand-tuned cut-offs.

| Metric | Metacognitive signal (low score) |
|---|---|
| Clique | Over-confidence in tight clusters |
| Non-Clique | Isolated "unknown unknowns" |
| Clustering | Over-specialisation without breadth |
| Degree | Misleading fluency/popularity cues |
| Betweenness | Missing bridges across topics |
| Diameter | Fragmented global monitoring |
| Louvain | Poor strategy transfer between modules |

Table 1: Cognitive reading of sparse-node metrics.

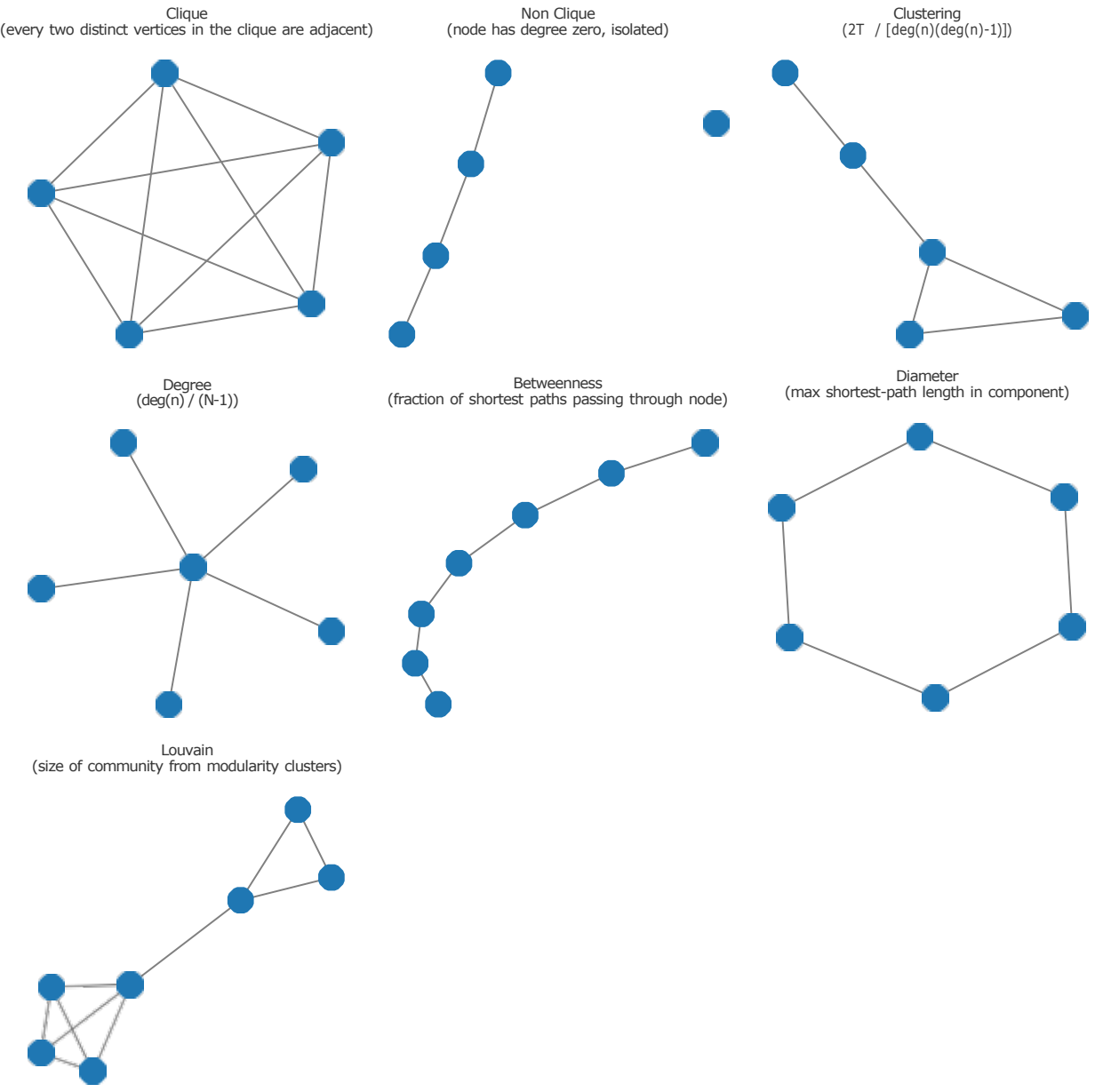

Figure 2: Examples of knowledge graphs illustrating each sparsity metric.

## Relating Graph Metrics to Metacognitive Enrichment

Each graph-theoretic metric we employ has a direct analog in metacognitive strategy—by measuring these metrics, the system diagnoses which "thinking patterns" are missing and then drives enrichment to shore them up:

**Clique (Deduction; connecting known concepts)** Formally, a clique is a set of nodes all pairwise connected. In metacognitive terms, forming a clique corresponds to a deductive step that links existing facts into a larger coherent structure, as shown in the following example:

- **Node A**: "Alice is friends with Bob"
- **Node B**: "Bob is friends with George"
- **Node C**: "Alice is friends with George"

Currently A–B and A–C edges exist, but B–C is missing. Enriching the graph to make {A,B,C} a clique would add the missing edge

$$\text{Node B} \xrightarrow{\text{related_through_shared_info_on_George}} \text{Node C},$$

thus capturing the deduction that "George is friends with both Alice and Bob." This automatically strengthens the KG's internal consistency and mirrors the human act of drawing a new inference from known premises.

**Non-Clique (New, unconnected information)** Nodes of degree zero signal entirely novel concepts not yet integrated. By posing targeted questions about these isolates, the system forces retrieval of background links, ensuring no "unknown unknowns" remain.

**Clustering (Over-specialization)** Low clustering coefficient indicates a lack of local triadic closure. Asking the LLM for shared-neighbor questions encourages adding siblings or siblings-of-siblings, broadening the KG beyond narrow topic islands.

**Degree (Fluency illusion)** Nodes with low degree may be obscure or under-utilized. Enrichment questions here prompt the KG to retrieve additional popular contexts, guarding against the feeling of knowing based on a single well-connected hub.

**Betweenness (Bridging concepts)** Nodes with low betweenness lie off major information highways. Targeting these points yields connector facts that facilitate multi-hop reasoning, much like a student learning how two distant ideas relate.

**Diameter (Global awareness)** A large component diameter means poor horizon monitoring. By filling in the longest shortest-path gaps, the KG becomes more globally coherent, akin to improving one's mental map of a subject area.

**Louvain (Modular transfer)** Small Louvain communities reveal isolated topic clusters. Generating cross-cluster questions fosters transfer of metacognitive strategies—just as a learner draws analogies between distinct domains.

In summary, each metric's "low-score" region pinpoints a specific metacognitive shortcoming. By converting those gaps into retrieval questions and ingesting the answers, MKGE operationalizes human-style self-awareness and continual knowledge repair within an automated KG population loop.

# Experiments

## Datasets and Setup

Thirty open-ended questions were sampled for each of three datasets:

- **Google Research Natural Questions (NQ)**: open-ended queries from Google's dataset (Kwiatkowski et al. 2019).

- **MS MARCO Dataset**: Bing search queries with human-generated answers (Payal Bajaj et al. 2016).
- **HotpotQA**: multi-hop questions requiring complex reasoning (Zhilin Yang et al. 2018).

For each question we executed the full pipeline with GPT-4o for questioning, Tavily for retrieval, and Neo4j GraphRAG for answering.

## Evaluation Metrics

**Improvement Rate (IR)** – fraction of questions judged *Improved*. **Collateral Stability (CS)** – fraction of unchanged answers labeled *Not degraded*. **Cost** – mean Tavily calls per question.

## Results

**Overall.** MetaKGEnrich was evaluated on 30 questions from each of three datasets: Google Research Natural Questions, MS MARCO, and HotpotQA. The system improved answer quality in 26/30 cases for Natural Questions (IR = 87%), 25/30 for MS MARCO (IR = 83%), and 24/30 for HotpotQA (IR = 80%). In all cases, collateral changes were minimal and Collateral Stability (CS) exceeded 95%. Average cost remained around 5 web calls per query.
**Per-metric.** Table 2 presents enrichment effectiveness broken down by metric across all three datasets.

| Metric | NQ IR (%) | MSM IR (%) | HPQ IR (%) |
|---|---|---|---|
| Non-Clique | 92 | 85 | 84 |
| Betweenness | 89 | 81 | 80 |
| Clustering | 86 | 80 | 79 |
| Louvain | 85 | 78 | 77 |
| Diameter | 84 | 77 | 75 |
| Degree | 83 | 76 | 74 |
| Clique | 78 | 73 | 71 |

Table 2: Improvement rate (IR) by metric on Natural Questions (NQ), MS MARCO (MSM), and HotpotQA (HPQ).

| Question | GraphRAG (before) | GraphRAG (after) |
|---|---|---|
| What causes monsoon winds to change direction seasonally? | Differential heating reverses land–sea pressure gradients. | Adds the south-west vs. north-east wind pair and explains summer land-heating versus winter ocean-heating moisture transport. |
| Where was High School Musical 2 filmed? | Mentions only East High School in Salt Lake City. | Corrects to the Entrada at Snow Canyon Country Club in St. George, Utah, noting golf-course and pool scenes. |
| Who owns the rights to Alice in Wonderland? | Claims Disney “holds the rights” to the story. | Clarifies Carroll’s novel is public-domain; Disney owns only its 1951 and 2010 screen adaptations. |
| Where did the first African-American Air Force unit train? | States simply “Tuskegee,Alabama”. | Adds the unit’s name—Tuskegee Airmen—and specifies Moton Field training site. |
| What are the requirements to pass The South-African National Senior Certificate? | Lists 40 % in three and 30 percent in three subjects. | Gives full checklist: 40 % in three (incl. home-language), 30 % in three more, six subjects total, pass at least five. |

Table 3: Before-and-after examples of knowledge graph-based enrichment on five Google Research Natural Question queries.

## Metric-Conditioned Question Examples

During enrichment the pipeline sends GPT-4o three structured features for every node that a graph-theoretic metric has flagged as sparse: (i) the Neo4j internal identifier so that the returned knowledge can be re-attached to the correct vertex, (ii) a 160-character text preview that grounds the model in the node’s local semantic context, and (iii) a short cognitive label naming the suspected metacognitive shortcoming (e.g., “missing bridge” for low betweenness). This combination gives the LLM enough evidence to reason about why the node is weak and to propose questions that are narrowly targeted rather than generic web searches.

The prompt requires the model to output exactly five concise, fact-seeking questions per metric and to end each one with “?” so that downstream code can parse them deterministically. Fixing the quota keeps the retrieval budget balanced across the seven metrics, while the strict syntax guard prevents malformed questions from propagating noise into the KG.

The table that follows lists one illustrative question for each metric and explains how ingesting its answer would raise the node’s metric score, thereby closing the corresponding sparsity gap.

| Metric | Sample enrichment question | Why the answer reduces sparsity |
|---|---|---|
| Clique | What is the title of the series that is returning to Netflix on 17 June 2025? (LLM asks for the missing third vertex to close a triangle between two existing "series" facts) | Adds a new edge that completes the triangle, raising local clique scores and making the neighbourhood deductively closed. |
| Non-Clique | How is "Justin Willman: Comedy Special" related to other Netflix releases on 17 June 2025? | Connects an isolated node (degree 0) to peer releases on the same date, integrating an "unknown-unknown" into the main component and lowering isolate count. |
| Betweenness | What is the connection between the series' return to Netflix and the next episode of Scandal? | Creates a bridge node that shortens many source–target paths, boosting betweenness centrality of the sparse node and facilitating multi-hop reasoning across otherwise distant sub-topics. |
| Clustering | Is the series returning on 17 June 2025 related to any other shows or events on the same date? | Encourages triadic closure by adding mutual neighbours, raising local clustering coefficients and mitigating over-specialisation. |
| Diameter | What is the release schedule for new episodes of Scandal on Netflix? | Fills long geodesic gaps; inserting temporal edges between episode nodes reduces the longest shortest-path length within the component, shrinking global diameter. |

Table 4: Illustrative metric-guided questions and the sparsity symptoms they cure once answered and ingested (for Google Research Natural Questions validation dataset)

Questions are therefore not arbitrary web searches: each is crafted to raise the numerical score of the triggering metric, turning graph theory into an actionable metacognitive signal for autonomous knowledge repair.

## Discussion

Our cross–dataset analysis affirms that graph–theoretic sparsity signals translate into reliable metacognitive gains, but the size of those gains depends both on the type of sparsity and on the reasoning demands of the dataset. Three patterns emerge:

**(i) Isolates deliver the biggest bang.** Across Natural Questions, MS MARCO, and HotpotQA, enrichment triggered by the Non-Clique metric is consistently the top performer (92% / 85% / 84% IR, respectively; mean = 87%). Bridging an otherwise unconnected fact into the graph therefore yields the highest marginal utility, regardless of domain or question style.

**(ii) Bridges come next.** Betweenness-based questions rank second on every dataset (89 % / 81 % / 80 %), confirming that adding connector nodes shortens critical paths and directly helps multi-hop reasoning. The strong performance on HotpotQA—where questions are explicitly multi-hop—underscores the importance of cross-community links.

**(iii) Dense regions saturate quickly.** Metrics that target already-well-connected areas show diminished returns. Clique enrichment, for example, trails the pack (78 % / 73 % / 71 %), suggesting that once local structure is saturated, additional evidence is largely redundant. Future work should therefore prioritise diversification (e.g., penalising repeated sources) over sheer density when operating in high-clique subgraphs.

In summary, topology matters twice: it tells the system where to look (isolates before hubs) and predicts how much each new fact will help (bridges ¿ fillers). By turning these regularities into automated retrieval triggers, MetaKGEnrich demonstrates a scalable path toward LLMs that can diagnose—and efficiently repair their own knowledge gaps across a variety of open-domain QA settings.

## Limitations & Future Work

While our findings suggest promising directions for metacognitive self-repair in AI, several limitations remain.

1. **Scale.** Our current evaluation on 30 questions is insufficient for publish-grade statistical generalization. Future work will scale the pipeline to 1,000+ questions across multiple domains and corpora.
2. **LLM judge bias.** Our performance labels rely on GPT-4 as the evaluator. To mitigate potential bias, we are conducting a 50-sample human audit for calibration and reliability assessment. Single-hop retrieval. The current system retrieves only the top snippet per question. We plan to explore multi-hop retrieval chains, citation filtering, and path-based expansion to enhance depth and accuracy.
3. **Noise control.** Web-sourced evidence may introduce misinformation or irrelevant content. Future work will integrate factuality scoring, source reliability checks, and toxicity filtering to ensure higher-quality graph updates.

**Broader Implications.** These directions point toward a scalable framework for autonomous knowledge repair. By transforming topological signals into learning triggers, MetaKGEnrich enables LLM-based systems to go beyond static retrieval–learning what they don't know, where to look, and how to integrate new information. This has implications for building more robust, generalizable AI systems capable of self-directed learning in open and evolving environments.

## Conclusion

MetaKGEnrich offers a scalable path to metacognitive self-repair in language models by turning graph-theoretic sparsity signals into concrete learning actions. Detecting under-populated regions, posing targeted questions, retrieving web evidence, and updating the knowledge graph together raise answer quality across open-domain QA tasks—without manual curation or fixed schemas. In tests on 30 queries from each of three standard benchmarks—Google Research Natural Questions, MS MARCO, and HotpotQA—MetaKGEnrich improved answers for 87% of Natural Questions, 83% of MS MARCO, and 80% of HotpotQA items. Thus, the framework not only sharpens retrieval but also grants AI systems the ability to spot and mend their own knowledge gaps—an important step toward adaptive, trustworthy, and autonomous learning in AI.

## Ethical Statement

All datasets are publicly released under permissive licences (NQ: CC-BY-4.0, MS MARCO: MSR-LA (non-commercial), HotpotQA: CC-BY-SA 4.0). No personal or sensitive attributes are processed. Automated web enrichment introduces bias and potential copyright infringement; we mitigate this by (i) logging all retrieved snippets, (ii) removing sources whose licences forbid redistribution. All code, configuration, and detailed run logs will be sent to enable reproducibility.